\pdfoutput=1

\documentclass[11pt]{article}

\usepackage[]{acl}

\usepackage{times}
\usepackage{latexsym}

\usepackage[T1]{fontenc}

\usepackage[utf8]{inputenc}

\usepackage{microtype}
\usepackage{multirow}
\usepackage{booktabs}
\usepackage{adjustbox}
\usepackage{amsmath}

\newcommand{\Modelsp}{TRT }

\title{Leveraging Knowledge in Multilingual Commonsense Reasoning}

\author{Yuwei Fang, Shuohang Wang, Yichong Xu, \\
\bf Ruochen Xu, Siqi Sun, Chenguang Zhu, Michael Zeng \\
Microsoft Cognitive Services Research Group \\
\small \{\tt yuwfan, shuowa, yicxu, ruox, siqi.sun, chezhu, nzeng\}@microsoft.com}

\begin{document}
\maketitle
\begin{abstract}

Commonsense reasoning (CSR) requires the model to be equipped with general world knowledge.
While CSR is a language-agnostic process, most comprehensive knowledge sources are in few popular languages, especially English.
Thus, it remains unclear how to effectively conduct multilingual commonsense reasoning (XCSR) for various languages.
In this work, we propose to utilize English knowledge sources via a translate-retrieve-translate (TRT) strategy.
For multilingual commonsense questions and choices, we collect related knowledge via translation and retrieval from the knowledge sources. The retrieved knowledge is then translated into the target language and integrated into a pre-trained multilingual language model via visible knowledge attention.
Then we utilize a diverse of 4 English knowledge sources to provide more comprehensive coverage of knowledge in different formats.
Extensive results on the XCSR benchmark demonstrate that TRT with external knowledge can significantly improve multilingual commonsense reasoning in both zero-shot and translate-train settings, outperforming 3.3 and 3.6 points over the previous state-of-the-art on XCSR benchmark datasets (X-CSQA and X-CODAH).
\end{abstract}

\section{Introduction}
Commonsense reasoning (CSR) is one of the key challenges in natural language understanding.
It requires a model to integrate world knowledge into language modeling to produce answers.
A large number of tasks have been proposed to evaluate commonsense reasoning in English, such as COPA~\cite{copa} and CSQA~\cite{csqa}. 

Most recently, multilingual commonsense reasoning (XCSR) begins to draw attention from the community and a number of datasets emerged, e.g., X-CSQA~\cite{xcsr}, X-CODAH~\cite{xcsr}, XCOPA~\cite{xcopa}. The goal of XCSR is to extend a model's commonsense capability beyond language barriers.

\begin{figure}[t!]
\centering
\includegraphics[width=85mm]{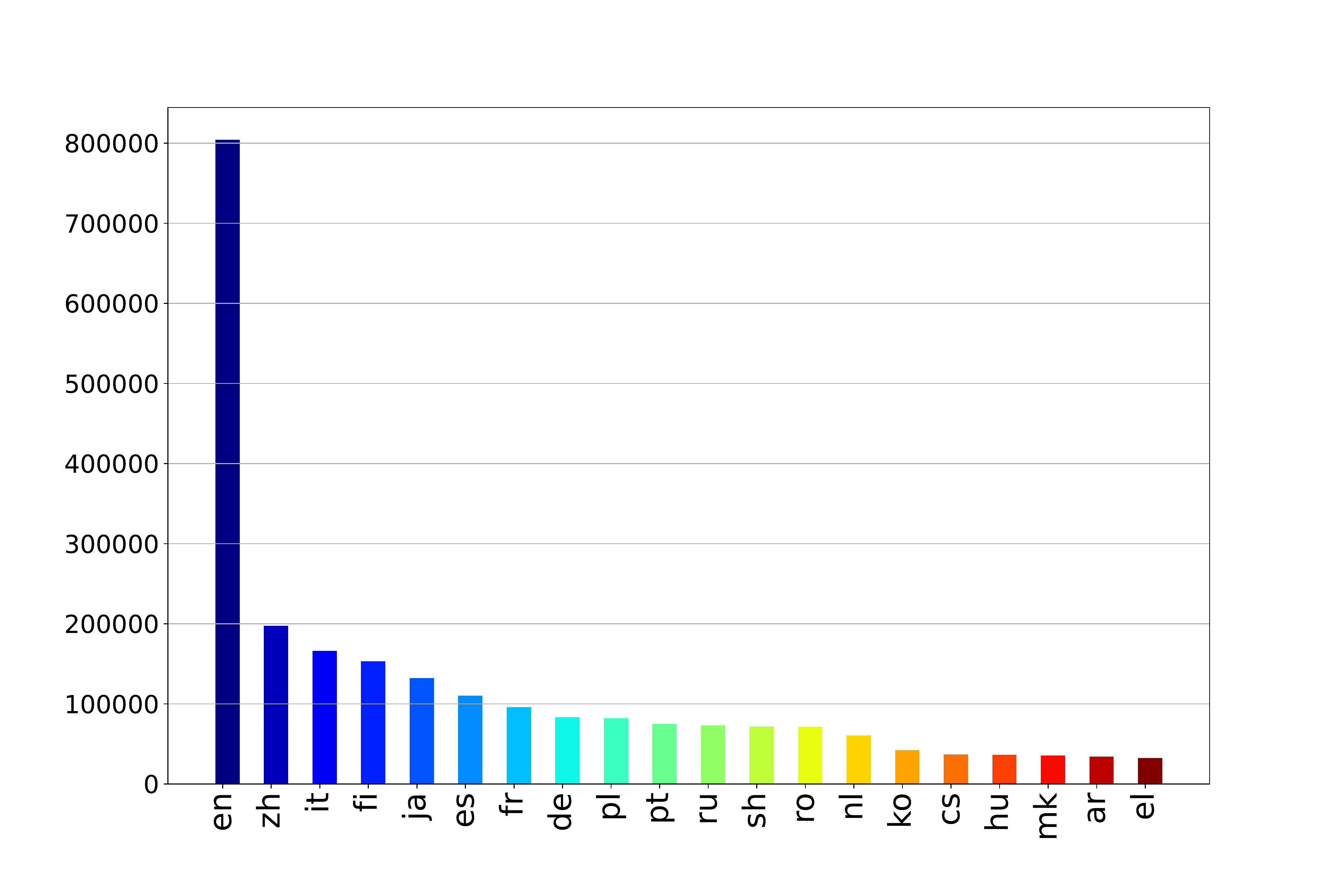}
\caption{\label{fig:lang_def}
Number of total definitions per language. The statistics are generated from Wiktionary 2021-10-01 dump. 
There are 55 languages with 10,000 or more definitions and we list top 20 languages by the definitions count here.
}
\vspace{-15pt}
\end{figure}

\begin{figure*}[ht]
\centering
\resizebox{0.95\textwidth}{!}{%
\includegraphics{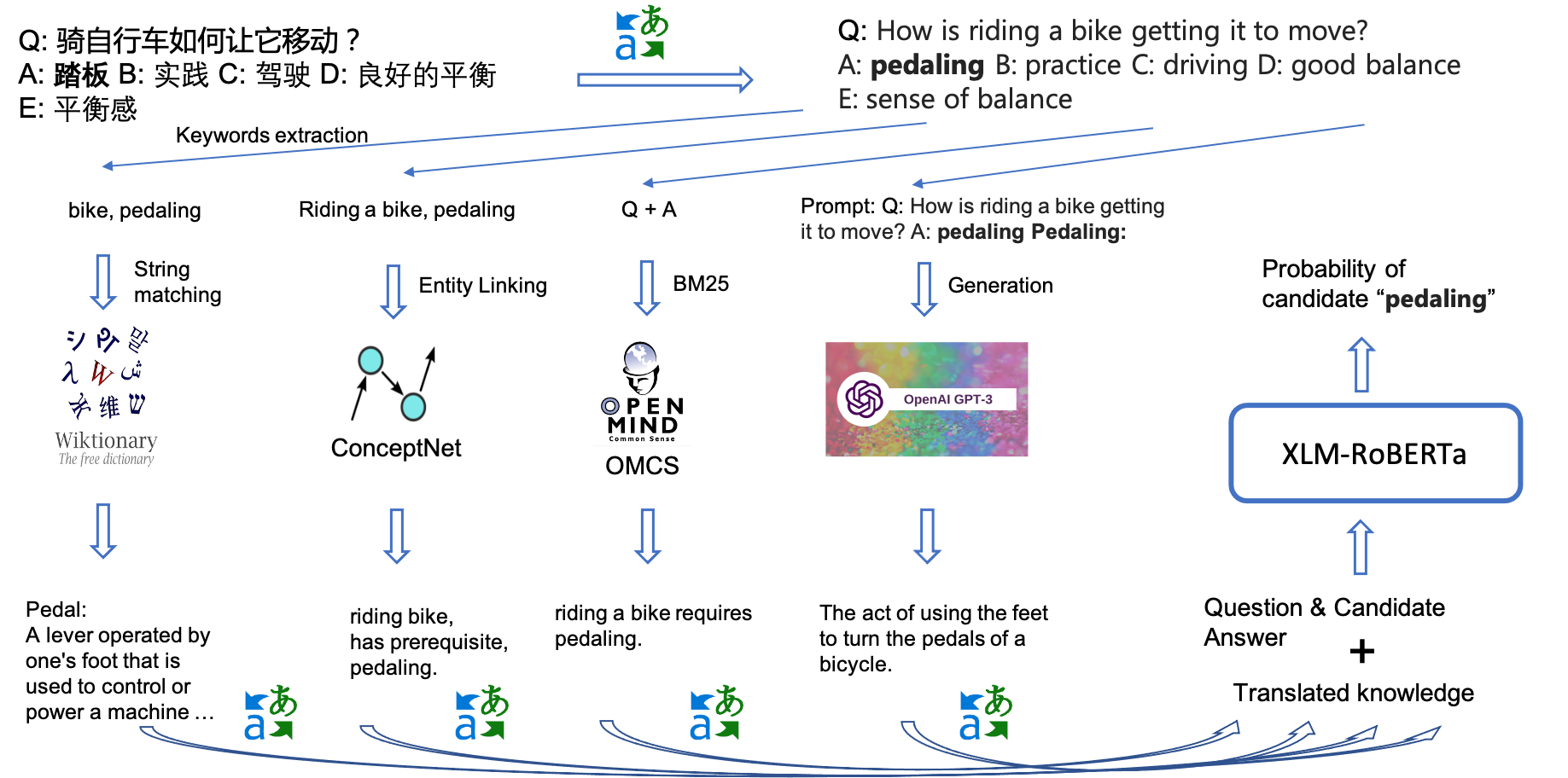}
}
\caption{An overview of our framework for multilingual commonsense reasoning. Given the question and candidate answers in the target language (Chinese), we first translate it into English, then retrieve related knowledge from four English knowledge sources and translate the retrieved knowledge back into the target language. The retrieved knowledge, along with question and candidate answer, are fed into the multilingual pretrained language model for answer prediction.
\label{fig:overview}
}
\end{figure*}

To solve commonsense reasoning tasks, it is essential to fuse human created knowledge into pre-trained language model (PLM)~\citep{lin2019kagnet,feng2020scalable,yu2020survey,dekcor}.
For example, DEKCOR~\cite{dekcor} integrates knowledge from ConceptNet~\cite{conceptnet} and Wiktionary~\footnote{https://www.wiktionary.org/} into the ALBERT model \cite{Lan2020ALBERT:} for commonsense question answering. However, most existing knowledge sources are crafted in a few popular languages, especially English. 
For example, Figure~\ref{fig:lang_def} shows the number of total definitions in English is much more than any other languages based on the statistics from Wiktionary 2021-10-01 dump.
Thus, it remains an open question how to tackle XCSR with a lack of curated knowledge in the target language.





In this paper, we propose a translate-retrieve-translate (TRT) solution to utilize English knowledge sources for XCSR. 
Specifically, given a commonsense reasoning question (possibly concatenated with a candidate answer) in the target language, we first translate it into English. 
Next, we retrieve related knowledge from English knowledge sources.
The retrieved knowledge is then translated back into the target language.
Finally, the knowledge is integrated into a multilingual language model via visible knowledge attention mechanism to answer the question.

Another contribution of our work is that we utilize a diverse set of 4 English knowledge sources to provide a more comprehensive coverage of knowledge in different formats. Specifically, we utilize unstructured text corpus (Open Mind Common Sense \cite{omcs}), structural knowledge graph (ConceptNet \cite{conceptnet}), dictionary (Wiktionary) and large-scale language model (GPT-3 \cite{gpt3}). 
Given an input query, we utilize information retrieval, entity linking, and model inference to obtain knowledge from corresponding sources.

We conduct extensive evaluation of our model on the  multilingual commonsense reasoning benchmark X-CSQA and X-CODAH~\cite{xcsr}. The results demonstrate the effectiveness of our proposed translate-retrieve-translate solution with multiple knowledge sources. For example, in the zero-shot transfer setting, TRT with Wiktionary can improve 1.9 and 2.7 points over the baselines. For translate-train setting, TRT with Wiktionary and OMCS outperform 1.6 and 1.0 over the baselines.

We summarize the main contributions of this work as follows. $(i)$ We propose a translate-retrieve-translate (TRT) solution to utilize English knowledge sources for multilingual commonsense reasoning. $(ii)$ We comprehensively explore four knowledge sources in different formats and prove their helpfulness for both X-CSQA and X-CODAH.
$(iii)$ We achieve the first place on XCSR leaderboard, outperforming 3.3 and 3.6 points over the previous state-of-the-art works.

\begin{table*}[t!]
\centering
\small
\begin{tabular}{lcccc}
\toprule
Knowledge Source & Knowledge Format & Query Format & Retrieved Knowledge & Retrieval Method  \\
\midrule
Wiktionary & Dictionary & Content Word & Definition & String Matching   \\
ConceptNet & Entity-Relation Triplets & Entity Pair & Entity-Relation Triplet & Entity linking    \\
OMCS & Text in Sentences &  Sentences &  Sentences & BM25  \\
GPT-3 & Parameters  & Unstructured Text & Unstructured Text & Conditional Generation \\
\bottomrule
\end{tabular}
\caption{Different knowledge resources for retrieval.}
\label{tbl:kg_sources}
\end{table*}

\section{Related Work}

\paragraph{Multilingual Commonsense Reasoning }
Model ability of commonsense reasoning has been widely explored by multiple downstream tasks. In early works, Winograd schema challenge~\cite{levesque2012winograd} is to disambiguate the reference of a pronoun~\cite{levesque2012winograd} and Choice of Plausible Alternatives (COPA)~\cite{roemmele2011choice} is to select cause or result for a premise. Later on, larger scale datasets, such as SWAG~\cite{zellers2018swag}, CODAH~\cite{codah}, and CommonsenseQA~\cite{csqa}, have been constructed for commonsense knowledge learning. Recently,  commonsense reasoning tasks have been extended to multilingual setting, such as X-CSQA~\cite{xcsr}, X-CODAH~\cite{xcsr}, XCOPA~\cite{xcopa}. In paper, we focus on training model to learn commonsense knowledge in multiple languages.

\paragraph{External Knowledge Fusion}
Knowledge bases are the most important external sources to help models learn the ability of commonsense reasoning. A wide range of knowledge resources, such as ConceptNet~\cite{conceptnet}, Wikipedia, Freebase~\cite{pellissier2016freebase}, and some KBs in domain~\cite{fader2011identifying}, can be fused into the model. 
\citet{lobue2011types} explored how commonsense knowledge involved in recognizing textual entailments.
\citet{guan2020knowledge} utilize commonsense knowledge to generate reasonable stories. 
\citet{bi2019incorporating} incorporate external Knowledge into question answering. 
\citet{dekcor} fuse the ConceptNet~\cite{conceptnet} and Wikionary into the model for solving CommonsenseQA.
In this paper, we will follow this direction and have a wider exploration of leveraging different sources for multiligual commonsense reasoning. 

\section{Approach}

In this section, we first formalize the multilingual commonsense reasoning (XCSR) task (Section~\ref{sec:prob}). Then we describe more details about our commonsense knowledge resources (Section~\ref{sec:kg_sources}).
Next, we introduce our proposed translate-retrieve-translate (TRT) solution to obtain the multilingual knowledge (Section ~\ref{sec:obtain}).
 Finally, we introduce how to fuse the obtained knowledge into multilingual pre-trained language models by employing the visible attention mechanism (Section~\ref{sec:fusing}). The overview of the framework is illustrated in Figure~\ref{fig:overview}.

\subsection{Problem Formulation}
\label{sec:prob}
We denote a language by $l \in L$, where $L = \{en, fr, de, zh, \cdots \}$. 
Given a commonsense question $q^l$ in the target language $l$, the goal is to choose the correct answer from $N$ candidates $\{c_{1}^{l}, c_{2}^{l}, \cdots, c_{N}^{l}\}$.
We assume there are one or more external knowledge sources to provide world knowledge in various formats for commonsense reasoning.
Each time the model retrieves knowledge using the question-candidate pair as query, i.e., $p^l = [q^l, c_{i}^{l}]$.

\subsection{Commonsense Knowledge}
\label{sec:kg_sources}
Commonsense knowledge 
are critical to the performance of a commonsense reasoning (CSR) model. 
Previous methods for CSR primarily integrate knowledge from one or two sources~\cite{dekcor}. In this work, we conduct comprehensive experiments by leveraging commonsense knowledge from 4 different resources: unstructured text corpus (Open Mind Common Sense),  knowledge graph (KG) (ConceptNet), dictionary (Wiktionary), and pre-trained language model (PLM) (GPT-3). 
Open Mind Common Sense (OMCS)~\cite{omcs} is a large commonsense knowledge base which has accumulated millions of facts. ConceptNet~\cite{conceptnet} is a semantic network built on top of MOCS. Wiktionary provides the definitions for all the words. GPT-3~\cite{gpt3} is a large-scale pre-trained language model to generate knowledge by feeding a query. These knowledge resources are saved in quite diverse formats as the analysis shown in Table~\ref{tbl:kg_sources}. To retrieve the knowledge, we will consider different query formats and retrieval methods in the next section.







\subsection{Knowledge Retrieval}
\label{sec:obtain}
Most large-scale knowledge sources in either academia or industry are crafted in a few popular languages, especially in English (see Figure~\ref{fig:lang_def} as an example). 
To obtain knowledge for low-resource languages, we propose a translate-retrieve-translate (TRT) solution.
In detail, we first use a machine translation tool to translate the query in all  languages into English.
Then, we can retrieve knowledge from English knowledge sources using the translated query.
The retrieved knowledge can be then translated back into original languages for model training.

As a knowledge source usually contains vast amount of information, we need to retrieve and leverage only the related knowledge for a given query $p^{l}$. 
Next we introduce the details of knowledge retrieval for 4 knowledge sources.


\paragraph{Word definition retrieval from Wiktionary}
Every word has its own definition but not all of them are delivering knowledge for commonsense reasoning. In this work, we mainly focus on retrieving the content words, such as nouns, verbs, and adjectives, and the words harder to understand by multilingual language models. In detail,
after part-of-speech tagging of the sequence, we 
select the nouns, verbs and adjectives as the candidate words.
Then, we mask one word at a time and compute its masked language model (MLM) probability by pre-trained multilingual language model, XLM-RoBERTa~\cite{xlmr}.
We select top-N words with lowest MLM probability for dictionary retrieval.
If the original word is not in Wiktionary, we try to find its lemmazied form.
The first definition entry in Wiktionary is the retrieved knowledge.

\paragraph{Structured knowledge retrieval from ConceptNet}
A knowledge graph can provide relation information between entities. 
We enumerate pairs of candidate words from the input sequence and check whether there exists a relation between them in the knowledge graph ConceptNet.
If so, we retrieve the corresponding triplet as the external knowledge.

\paragraph{Unstructured text retrieval from OMCS}
Open Mind Common Sense (OMCS) consists of knowledge in natural language description. We first  build a search index \footnote{https://lucene.apache.org/pylucene/} for all the sentences in OMCS.   Then, whenever a new query comes, we retrieve the highest ranked sentence based on BM25 as the external knowledge text.

\paragraph{Knowledge Generation with GPT-3}
Previous research shows that large-scale PLM contains rich knowledge implicitly \cite{roberts-etal-2020-much, kassner-etal-2021-multilingual}.
Thus, we use one of the largest PLM, GPT-3 \cite{gpt3}, to generate related knowledge given the query.
As GPT-3 requires a prompt with input and output examples, we feed it with a few examples with a query and the knowledge in designated format.
For example, given the word and its definition along with the query, GPT-3 will generate its version of definition of a word it thinks important in the input query. For the prompt that is not in English, we translate the English prompt into the target language. 



\begin{table*}[t!]
\centering
\resizebox{\linewidth}{!}{
\small 
\begin{tabular}{l|c|cccccccccccccccc|c}
\toprule
Dataset & Model & en & de & it & es & fr & nl & ru & vi & zh & hi & pl & ar & ja & pt & sw & ur & avg \\
\midrule
\multirow{5}{*}{X-CSQA} & mBERT & 38.8 & 29.6 & 36.4 & 35.3 & 33.8 & 32.6 & 32.7 & 22.2 & 37.8 & 21.1 & 27.2 & 27.7 & 31.4 & 34.1 & 21.8 & 23.7 & 30.4 \\
& XLMR-B & 51.5 & 44.1 & 42.1 & 44.8 & 44.0 & 43.3 & 39.5 & 42.6 & 40.6 & 34.6 & 40.2 & 38.4 & 37.5 & 43.4 & 29.6 & 33.0 & 40.6 \\
& XLMR-L                     & 66.7 & 56.1 & 58.2 & 59.5 & 60.3 & 56.8 & 52.1 & 51.4 & 52.7 & 48.7 & 53.9 & 48.4 & 50.0 & 59.9 & 41.6 & 45.2 & 53.8 \\
& MCP (RL)              & 69.5 & 59.3 & 60.3 & 61.4 & 60.0 & 61.1 & 57.5 & 55.7 & 56.7 & 51.3 & 56.1 & 52.3 & 50.2 & 60.7 & 43.3 & 48.8 & 56.5 \\
\cmidrule{2-19}
& \Modelsp              & \textbf{71.0}	& \textbf{61.2} & \textbf{63.0} & \textbf{65.1} & \textbf{65.1} & \textbf{62.8} & \textbf{57.8} & \textbf{58.9} & \textbf{56.3} & \textbf{56.1} & \textbf{59.4} & \textbf{56.2} & \textbf{54.7} & \textbf{64.6} & \textbf{51.0} & \textbf{53.9} & \textbf{59.8} \\
\midrule \midrule
\multirow{5}{*}{X-CODAH} & mBERT & 42.9 & 33.1 & 33.5 & 33.8 & 35.2 & 33.7 & 31.9 & 22.8 & 38.0 & 26.5 & 31.0 & 34.8 & 34.0 & 37.2 & 30.8 & 31.5 & 33.2 \\
& XLMR-B & 50.1 & 45.8 & 44.4 & 44.2 & 45.2 & 42.0 & 44.1 & 43.2 & 44.6 & 38.1 & 41.9 & 37.8 & 42.0 & 44.1 & 35.6 & 34.6 & 42.4 \\
& XLMR-L                     & 66.4 & 59.6 & 59.9 & 60.9 & 60.1 & 59.3 & 56.3 & 57.4 & 57.3 & 49.1 & 57.5 & 51.2 & 53.8 & 58.2 & 42.2 & 46.6 & 56.0 \\
& MCP (RL)               & \textbf{69.9} & 60.7 & 61.9 & 60.7 & 61.4 & 60.7 & 58.6 & 62.3 & 61.9 & 53.7 & 59.0 & 54.1 & 54.7 & 60.8 & 44.6 & 48.0 & 58.3 \\
\cmidrule{2-19}
& \Modelsp              & 69.1	& \textbf{65.3}	& \textbf{62.5}	& \textbf{64.4}	& \textbf{64.3}	& \textbf{64.5}	& \textbf{61.8}	& \textbf{64.6} &	\textbf{63.3} & \textbf{57.1} & \textbf{62.7} & \textbf{57.6} & \textbf{61.6} & \textbf{64.3} & \textbf{52.5} & \textbf{55.1} & \textbf{61.9} \\

\bottomrule
\end{tabular}
}
\caption{
 Overall test results on the multilingual commonsense reasoning benchmark XCSR. Results of mBERT~\cite{mbert}, XLMR-B, XLMR-R~\cite{xlmr}, MCP(RL)~\cite{xcsr} for X-CSQA and X-CODAH are from XCSR leaderboard~\cite{xcsr}.
We submit the test prediction with the best dev result in table~\ref{tbl:xcsr_full_dev} to the XCSR leaderboard for evaluation. Leaderboard: \href{https://inklab.usc.edu//XCSR/leaderboard}{https://inklab.usc.edu//XCSR/leaderboard}
}
\label{tbl:xcsr}
\end{table*}

\subsection{Fusing Knowledge into Multilingual Language Model}
\label{sec:fusing}
Given the question answer pair $p^l = [q^l, c_{i}^{l}]$, we use the retrieval techniques to collect $K$ pieces of retrieved knowledge text: $S = [s_{1}, \cdots, s_{K}]$.

The most intuitive way is to concatenate them with $p^l$ as input to the multilingual pre-trained language model (XPLM) for answer generation, i.e., the input would be $I=$ [CLS] $q^l$  $c_{i}^{l}$ [SEP] $s_1$ [SEP] $\cdots$ $s_K$ [SEP].

However, this simple way may divert the original meaning of $p^l$ because of the introduced noise by appending $S$, as pointed out by~\citet{liu2020k,xu2021does}.
To remedy this issue, we adopt the visibility matrix~\cite{liu2020k, xu2021does} to limit the impact of knowledge set $S$ on the original question-candidate pair $p_l$.  
Specifically, in each transformer layer of XPLM, an attention mask matrix $M$ is added to the self-attention weights before softmax.

Suppose $t_j$ and $t_k$ are the $j$-th and $k$-th tokens from the input $I$. We set $M_{jk}$ to zero to allow attention from $t_j$ to $t_k$, and set $M_{jk}$ to $-\infty$ to forbid attention. 
$M_{jk}$ is set to zero if: i) both tokens belong to the input $p_l$, or ii) both tokens belong to the same knowledge $s_i$, or iii) $t_j$ is the token at the start position of linked word in $p_l$ and $t_k$ is from its correspond knowledge text. 
More formally, the mask matrix $M$ is
\begin{equation}
M_{jk} = 
    \begin{cases}
        0 & t_j, t_k \in p^l \\
        0 & t_j, t_k \in s_i \\
        0 & t_j \in p^l, t_k \in s_i  \\
        -\infty & \text{otherwise}
    \end{cases}
\end{equation}

For model training, let $z_0 \in R^{d}$, the \texttt{[CLS]} hidden state from the last layer, denotes the representation of encoding the question, candidate, and the corresponding retrieved knowledge. $d$ is the dimension of the output vector of the encoder. Then we calculate the prediction score $\hat{y}_i$ for each candidate $c_{i}^{l}$ with one linear layer, $\hat{y}_i = W_o z_0$, where $W_o \in R^{1 * d}$, followed by a softmax normalization upon all candidates, $\hat{y} = softmax([\hat{y}_i, \cdots, \hat{y}_N])$, where $N$ is the number of candidate for each question. The final loss function is the standard cross-entropy loss.

\section{Experiments}
\label{sec:exp}

In this section, we perform extensive experiments to explore the aforementioned \Modelsp solution with four knowledge sources on the multilingual commonsense reasoning benchmark XCSR~\cite{xcsr}.

\subsection{Datasets}
\begin{table}[t!]
\begin{adjustbox}{scale=0.95,center}
\small
\begin{tabular}{lcc}
\toprule

Dataset & X-CSQA & X-CODAH \\
\midrule
Task Format & QA & Scene Completion \\
\#Languages & 16 & 16 \\
\#Options & 5 & 4 \\
\#train & 8888 & 8476 \\
\#dev & 1000 & 300 \\
\#test & 1074 & 1000 \\
\bottomrule
\end{tabular}
\end{adjustbox}
\caption{Statistics of the two datasets in the multilingual commonsense reasoning benchmark XCSR}
\label{tbl:stat}
\end{table}
Table~\ref{tbl:stat} lists the statistics for the two datasets in XCSR. $(i)$ X-CSQA~\cite{xcsr} for commonsense question answering: given the human authored question that describes the relation between concepts from ConceptNet~\cite{conceptnet}, the model needs to choose the answer from five concepts. $(ii)$ X-CODAH~\cite{xcsr} for Scene Completion: given a prompt question and the subject of the subsequence sentence, the model needs to choose from four candidate complements that can be consistent with question in commonsense. 

\begin{table*}[t!]
\centering
\resizebox{\linewidth}{!}{
\small 
\begin{tabular}{l|c|cccccccccccccccc|c}
\toprule
Dataset & Model & en & de & it & es & fr & nl & ru & vi & zh & hi & pl & ar & ja & pt & sw & ur & avg \\
\midrule
\multicolumn{19}{l}{\textit{Zero-shot transfer (models are trained on English data) and evaluate on the target language}} \\
\midrule
\multirow{5}{*}{X-CSQA} & MCP (RL) & 69.0 & 57.6 & 57.2 & 57.9 & 59.9 & 56.1 & 55.2 & 56.0 & 56.6 & 48.8 & 56.4 & 52.5 & 50.8 & 58.3 & 42.5 & 47.4 & 55.1 \\
\cmidrule{2-19}
& \enspace + Wikt. & 70.7 & 59.5 & 60.2 & 61.4 & 59.5 & 58.5 & 56.6 & 55.6 & 58.3 & 51.2 & 56.0 & 55.6 & 52.0 & 60.6 & 46.8 & 49.1 & \textbf{57.0} \\
& \enspace + Cpnt. & 70.7 & 57.2 & 58.1 & 58.6 & 58.7 & 55.8 & 55.5 & 56.0 & 56.6 & 49.9 & 55.9 & 53.9 & 52.4 & 55.6 & 43.3 & 47.8 & 55.4 \\
& \enspace + OMCS & 70.5 & 59.9 & 59.3 & 60.5 & 60.0 & 56.8 & 55.3 & 56.1 & 57.3 & 48.9 & 56.4 & 53.4 & 51.6 & 59.0 & 46.7 & 48.0 & 56.2 \\
& \enspace + GPT-3 & 70.3 & 57.2 & 58.8 & 60.2 & 58.3 & 58.1 & 54.8 & 55.0 & 55.6 & 49.0 & 54.5 & 52.9 & 52.1 & 57.9 & 42.9 & 47.6 & 55.3 \\

\midrule
\multirow{5}{*}{X-CODAH} & MCP (RL) & 69.7 & 63.0 & 62.3 & 63.0 & 64.7 & 64.7 & 55.0 & 55.0 & 59.7 & 54.3 & 61.7 & 52.3 & 57.0 & 55.0 & 40.3 & 49.3 & 57.9 \\ 
\cmidrule{2-19}
& \enspace + Wikt. & 72.0 & 65.3 & 63.0 & 65.0 & 66.0 & 66.0 & 58.7 & 59.3 & 58.0 & 54.3 & 64.0 & 55.7 & 61.3 & 60.7 & 47.0 & 53.0 & \textbf{60.6} \\
& \enspace + Cpnt. & 72.3 & 68.3 & 65.7 & 65.0 & 66.0 & 64.3 & 60.3 & 57.0 & 58.3 & 55.0 & 65.3 & 53.7 & 57.3 & 59.7 & 46.3 & 52.0 & 60.4 \\ 
& \enspace + OMCS & 73.0 & 67.0 & 64.0 & 63.7 & 63.0 & 62.0 & 57.3 & 60.0 & 62.0 & 53.0 & 63.7 & 56.0 & 57.7 & 59.3 & 44.0 & 49.3 & 59.7 \\
& \enspace + GPT-3 & 71.7 & 62.0 & 64.3 & 62.3 & 65.0 & 62.3 & 56.7 & 55.3 & 58.0 & 54.3 & 64.7 & 55.0 & 59.3 & 60.0 & 42.7 & 52.7 & 59.1 \\
\midrule \midrule
\multicolumn{19}{l}{\textit{Translate-train (models are trained on English training data and its translated data) and evaluate on the target language}} \\ \midrule
\multirow{4}{*}{X-CSQA} & MCP (RL) & 69.4 & 59.3 & 60.6 & 60.9 & 60.8 & 57.9 & 57.0 & 58.2 & 58.0 & 50.4 & 58.3 & 55.1 & 53.9 & 60.3 & 47.1 & 50.9 & 57.4 \\
\cmidrule{2-19}
& \enspace + Wikt. & 70.0 & 61.7 & 61.2 & 61.1 & 60.9 & 59.8 & 59.8 & 59.3 & 59.6 & 53.8 & 59.7 & 58.1 & 54.3 & 60.5 & 51.8 & 52.8 & \textbf{59.0} \\
& \enspace + Cpnt. & 68.5 & 59.2 & 59.5 & 58.2 & 61.3 & 58.7 & 56.6 & 57.9 & 58.3 & 52.6 & 58.4 & 55.6 & 52.9 & 60.5 & 48.2 & 52.8 & 57.4 \\
& \enspace + OMCS & 71.7 & 61.1 & 63.6 & 62.8 & 60.3 & 58.6 & 58.1 & 59.3 & 58.5 & 51.7 & 58.1 & 56.1 & 54.2 & 60.4 & 48.6 & 53.4 & 58.5 \\
\midrule \midrule
\multirow{4}{*}{X-CODAH} & MCP (RL) & 71.0 & 70.7 & 66.3 & 69.7 & 70.7 & 66.7 & 63.7 & 62.3 & 62.3 & 60.3 & 64.7 & 59.3 & 59.7 & 67.7 & 57.0 & 57.7 & 64.4 \\
\cmidrule{2-19}
 & \enspace + Wikt. & 72.0 & 71.7 & 68.0 & 69.3 & 69.7 & 67.0 & 65.3 & 66.0 & 63.0 & 61.0 & 65.0 & 58.3 & 62.7 & 68.0 & 58.0 & 58.3 & 65.2 \\
& \enspace + Cpnt. & 70.7 & 68.7 & 67.0 & 68.0 & 68.0 & 68.3 & 65.0 & 62.0 & 61.7 & 56.3 & 65.0 & 61.7 & 62.3 & 66.3 & 60.0 & 57.3 & 64.3 \\
& \enspace + OMCS. & 74.7 & 69.7 & 67.3 & 67.7 & 67.7 & 68.3 & 62.7 & 65.3 & 65.3 & 58.7 & 68.3 & 62.0 & 64.0 & 68.3 & 56.7 & 59.7 & \textbf{65.4} \\
\bottomrule
\end{tabular}
}\caption{Comparisons for TRT with different knowledge sources in the zero-shot transfer and translate-train setting on the dev. Wikt. and Cpnt. are short for Wiktionary and ConceptNet. 
}
\label{tbl:xcsr_full_dev}
\end{table*}

\subsection{Baselines}
For X-CSQA and X-CODAH datasets, we mainly compare with the previous state-of-the-art MCP~\cite{xcsr} as well as other three multilingual pretrained langauge models: mBERT~\cite{mbert}, XLM-RoBERTa~\cite{xlmr} base and large models. MCP is based on XLM-RoBERTa model and further enhanced by intermediate fine-tuning on the multiple-choice question answering dataset MickeyProbe~\cite{xcsr}.


\begin{figure*}[!ht]
\centering
\includegraphics[width=140mm]{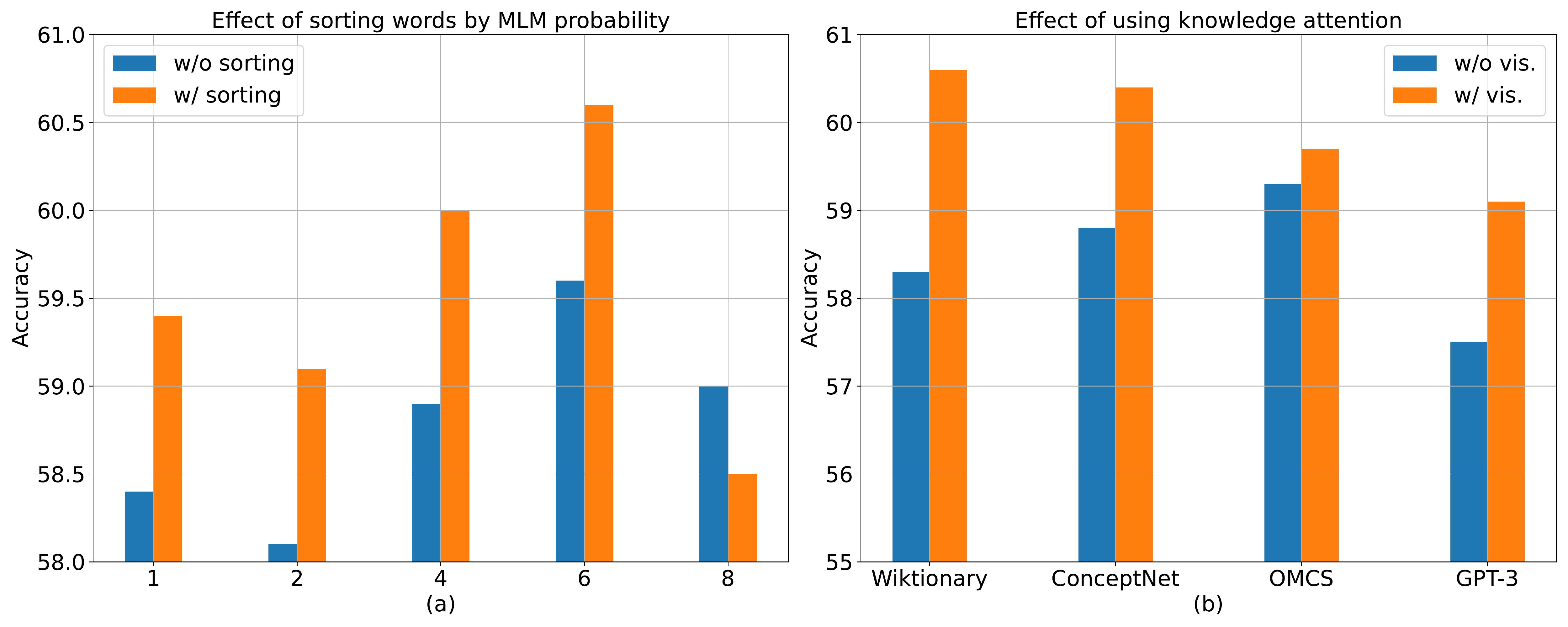}
\caption{
Effects of the number of word definitions and the visible attention mechanism on X-CODAH dataset.
Figure (a) shows the performance can be improved by increasing the number of definitions from 1 to 6.
Figure (b) shows visible attention can be helpful with all knowledge sources.
}
\vspace{-15pt}
\label{fig:analysis}
\end{figure*}

\subsection{Implementation Details}
We use Microsoft Machine Translator~\footnote{https://azure.microsoft.com/en-us/services/cognitive-services/translator/} for all translations, including translating the given query, the retrieved knowledge and English training data to other 15 languages. We will release these translations for academic usage. 
For Wiktionary, we use the dump of Wiktionary which includes 999,614 definitions.
We empirically obtaining 6 words definitions from Wiktionary for X-CODAH (see Figure~\ref{fig:analysis} (a)) and use the provided question concept and answer as two candidate words for X-CSQA.
For ConceptNet, we use ConceptNet version 5.7.0 \footnote{https://github.com/commonsense/conceptnet5}.
For GPT-3, we use the curie~\footnote{https://beta.openai.com/pricing} model.

Our model implementation is based on HuggingFace's Transformers Library~\cite{hf}. We conduct all experiments on 8 Nvidia V100-32GB GPU cards. We follow the configurations in XCSR to pretrain the MCP model based on XLM RoBERTa large except that the maximum sequence length is 256 and batch size is 32. The accuracy of the resulting MCP checkpoint on its dev set is 87.4. We then initialize with this checkpoint for further fine-tuning with different knowledge sources. During finetuning, we set the training epochs, batch size and gradient accumulation steps as 10, 4 and 2 respectively. 
The total batch size here is 64 by ``\textit{batch size per device × \# GPUs × \# gradient accumulation steps}''.
For hyper-parameter search, we sweep over the learning rates $\in \{1e-5, 3e-5, 5e-5, 3e-6, 5e-6\}$ and report the maximum results.

\subsection{Experimental Results}

\paragraph{Results on test set} Table~\ref{tbl:xcsr} summarizes our results on the hidden test set from XCSR leaderboard. \Modelsp outperforms all previous works by a significant margin on both datasets, achieving the average score of 59.8/63.7 with an absolute improvement of 3.3/3.6 over previous state-of-the-art MCP(RL).
For some low-resource languages, like Swedish, we observe even larger gains  with 7.7 and 7.9 improvements on X-CSQA and X-CODAH.

\paragraph{Effectiveness of different knowledge sources} Table~\ref{tbl:xcsr_full_dev} list the detailed comparisons among different knowledge sources in both zero-shot and translate-train setting. We observe the following findings from these results: $(i)$ Knowledge can be helpful for multilingual commonsense reasoning. For example, in the zero-shot setting, TRT with Wiktionary improve 1.9 and 2.7 points over the MCP baseline on X-CSQA and X-CODAH. In translate-train setting, there are 1.6 and 1.0 improvements.
$(ii)$ Wiktionary helps the most among all knowledge sources in both settings, except that OMCS performs slightly better than Wiktionary on X-CODAH in the translate setting. We hypothesize that the difficulty of understanding hardness words can be mitigated by incorporating additional knowledge as context. $(iii)$ The generated knowledge from GPT-3 can also improve over the baseline, without leveraging mahcine translation and explicit knowledge, which demonstrate the rich implicit knowledge in GPT-3. For example, for X-CODAH dataset, GPT-3 can outperform the baseline about 1.2 point. However, there still exist the gap between GPT-3 and designated knowledge format. We leave this one as future work to bridge the gap.

\paragraph{Effectiveness of sorting definitions by MLM probability}
In Section~\ref{sec:obtain}, we introduce using masked language model (MLM) to select the top-N hardness words with the lowest probability.
Therefore, we compare this strategy (w/ sorting) with randomly choosing the words. 
As shown in Figure~\ref{fig:analysis} (a), sorting by MLM probability can outperform the random selecting, especially with a smaller number of words, achieving the best performance with 6 words definitions.

\paragraph{Effectiveness of knowledge attention}
In Section~\ref{sec:fusing}, we mention that simply appending knowledge as additional context can be noise to some tasks like X-CODAH, a scene completion tasks.
Therefore, here we compare the model performance between full attention and visible knowledge attention on different knowledge sources.
As shown in Figure~\ref{fig:analysis} (b), knowledge attention (w/ vis.) can consistently outperform full attention (w/o vis.) on different knowledge sources. For example, there are 2.3 and 1.6 points improvement between them when integrating from Wiktionary and GPT-3.

\section{Conclusion}

In this work, we present the translate-retrieve-translate (TRT) strategy for multilingual commonsense reasoning that collects related knowledge via translation and then retrieval  from  the  knowledge  sources. We conduct extensive experiments by utilizing a diverse of four English knowledge sources, including Wiktionary, ConceptNet, OMCS and GPT-3. By using TRT with different knowledge sources, we achieve state-of-the-art results on XCSR leaderboard which demonstrates the effectiveness of our proposed methods. Future work includes more effective ways to incorporate the diverse knowledge sources into pre-training and fine-tuning stage for commonsense reasoning.

\bibliography{anthology,custom}
\bibliographystyle{acl_natbib}

\end{document}